\documentclass[journal]{IEEEtran}
\usepackage{colortbl}
\usepackage{cite}
\usepackage{graphicx}
\usepackage{amssymb}
\usepackage{xcolor}
\usepackage{url}
\definecolor{bestcolor}{RGB}{205,190,245}
\definecolor{secondcolor}{RGB}{232,225,250}
\usepackage{flushend}

\ifCLASSINFOpdf
\else
\fi
\usepackage{amsmath, amssymb, amsfonts, amsthm}
\begin{document}

\title{Memory-Native Non-Terrestrial Networks for Embodied Intelligence}
\author{Chengyang Li, Yikun Wang, Jiahui He, Yujie Wan, Shuai Wang, Yuan Wu,\\Yik-Chung Wu, Chengzhong Xu,~\emph{Fellow, IEEE}, and Huseyin Arslan,~\emph{Fellow, IEEE}
\vspace{-0.25in}
\thanks{

This work was supported in part by the CAS–TÜBİTAK Joint Call under the International Partnership Program of the Chinese Academy of Sciences (Grant No. 321GJHZ2025118M) and the Scientific and Technological Research Council of Türkiye (Grant No. 225N080), in part by the Science and Technology Development Fund of Macau SAR under Grant 0028/2025/AFJ, and in part by MOST and the Science and Technology Development Fund (FDCT) of the Macao SAR (File No. 0074/2025/AMJ).

Chengyang Li, Yikun Wang, and Yik-Chung Wu are with The University of Hong Kong, Hong Kong, China.

Jiahui He and Yujie Wan are with the Southern University of Science and Technology, Shenzhen, China, and also with the Shenzhen University of Advanced Technology, Shenzhen, China.

Shuai Wang is with the Shenzhen Institutes of Advanced Technology, Chinese Academy of Sciences, Shenzhen, China.

Yuan Wu and Chengzhong Xu are with the University of Macau, Macau SAR, China.

Huseyin Arslan is with the Istanbul Medipol University, Istanbul, Turkey.

Corresponding author: Shuai Wang ({\tt\footnotesize s.wang@siat.ac.cn}).
}
}

\maketitle

\begin{abstract}
Non-terrestrial networks (NTN) provide ubiquitous connectivity for embodied intelligence (EI), enabling robots in the wilderness to leverage cloud resources or report critical information to remote centers. However, the synergy is nontrivial due to the highly dynamic, resource-constrained, topology-varying, and task-oriented environment.
Existing memoryless NTN protocols become inefficient, since the decisions are driven by local channel conditions and instantaneous service demands.
To address these limitations, this paper proposes the memory-native NTN (MemNTN) paradigm that leverages long-horizon contexts for memory-augmented system optimization.
To realize this paradigm shift, we establish a dual-memory architecture that distinguishes between physical memory representing the state of the world and digital memory encoding historical network experience.
We develop memory acquisition, compression, valuation, update, and utilization mechanisms that facilitate cross-layer, memory-native decision-making, spanning from the physical and access layers up to the network and application layers.
Experiments in satellite embodied question answering (SEQA) demonstrate that the proposed MemNTN consistently outperforms conventional stateless NTN and terrestrial approaches.
\end{abstract}

\begin{IEEEkeywords}
Non-terrestrial networks, embodied intelligence, memory-native, satellite question answering.
\end{IEEEkeywords}

\IEEEpeerreviewmaketitle

\section{Introduction}

\begin{figure*}[!t]
    \centering
    \includegraphics[width=1.0\textwidth]{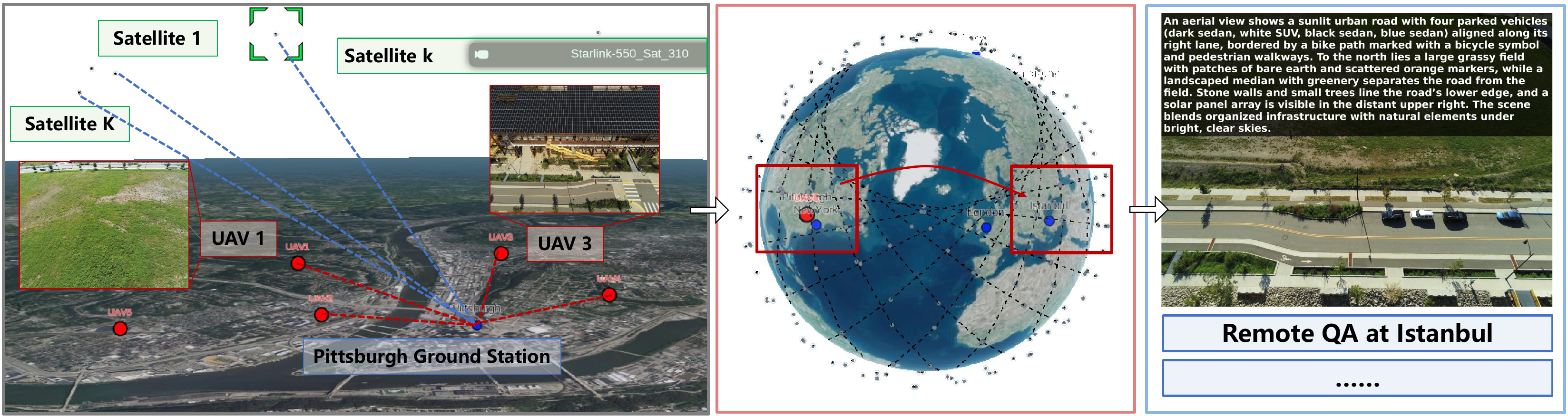}
    \caption{NTN enables embodied perception in Pittsburgh and remote question answering from Istanbul. Geographic data sources: OpenStreetMap, LEOPath, and Mega-NeRF/Mill-19.}
    \label{fig1}
\end{figure*}

Embodied intelligence (EI) plays a pivotal role in enhancing the safety and efficiency of emergency response, evolving from basic data collection to sophisticated search and rescue tasks~\cite{BravoArrabal2026Strengthening,Ghassemian2026Empowering}.
Realizing these EI capabilities depends heavily on the resilience of the underlying communication infrastructure, which remains vulnerable in wilderness and disaster-prone scenarios.
For instance, remote command centers require timely updates regarding real-time situational awareness \cite{Li2026MemoryCentric,Li2025Embodied}.
However, communication outages caused by damaged base stations or severe blockage may delay mission-critical decisions. Such disruptions not only degrade remote sensing and teleoperation performance \cite{Zheng2025Semantic}, but also hinder the deployment of cloud and edge computing~\cite{Zhai2024SECO}, limiting the overall timeliness and reliability of EI operations.

To address these challenges, non-terrestrial networks (NTN) with satellites and high-altitude platforms have become a promising solution~\cite{Jamshed2025Integrated, Jamshed2025NonTerrestrial,zhou2023aerospace, Azari2022Evolution, 2024Medina3GPP, Ma2024Satellite, Yin2020Integrated}. In contrast to terrestrial networks, NTN offers flexible communication links capable of covering remote areas in an on-demand manner. This capability ensures persistent connectivity for robotic operations, irrespective of terrestrial infrastructure limitations.
As depicted in Fig.~\ref{fig1}, the synergy between NTN and EI facilitates long-range semantic queries, where a user in Istanbul can retrieve spatio-temporal observations from robotic teams operating in Pittsburgh.

However, NTN-EI faces unique challenges spanning the physical layer (e.g., channel estimation), medium access control (MAC) layer (e.g., access protocols), network layer (e.g., mobility management), and application layer (e.g., task awareness), resulting in a highly dynamic, resource-constrained, topology-varying, and task-oriented environment. Existing NTN studies~\cite{Jamshed2025Integrated, Jamshed2025NonTerrestrial, zhou2023aerospace, Azari2022Evolution, 2024Medina3GPP, Ma2024Satellite, Yin2020Integrated} design routing, scheduling, and resource allocation based on local channel conditions and instantaneous service demands, which operate without
historical network experience and task-level context.
This makes them less effective in complex satellite embodied task environments.

To fill the gap, this paper proposes memory-native NTN (MemNTN), which enables all network nodes to function as intelligent agents that learn from historical experiences and make informed decisions.
The proposed MemNTN architecture introduces a dichotomous memory framework consisting of both physical memory and digital memory. Specifically, physical memory captures environmental geometry and semantic information (i.e., what the world looks like).
Digital memory records network experiences including channel state information (CSI), spectrum patterns, and scheduling records (i.e., what the network has experienced). Physical and digital memories are interconnected through a \emph{memory life cycle} that consists of memory acquisition, compression, valuation, update, and utilization mechanisms.
The core is the \emph{memory valuation} system that integrates the memory freshness, coverage, cost, reusability, and quality into a unified memory value function.
The memory values are then fed to a \emph{memory policy} system to drive memory-native cross-layer designs.

The contributions of this paper are fourfold:
\begin{enumerate}
\item We propose the MemNTN architectural framework, establishing the dichotomy between physical memory and digital memory.

\item We design multi-modal memory representations and memory life cycle mechanisms for building global memory across distributed NTN nodes.

\item We devise a memory value model that is disseminated throughout the vertical communication stack to derive a value-driven policy.

\item We conduct a case study to validate the MemNTN framework, demonstrating performance gains through high-fidelity simulations.
\end{enumerate}

\section{Memory-Native NTN}

\subsection{NTN for EI: Characteristics and Challenges}

NTN enables embodied applications in regions beyond the reach of terrestrial infrastructure.
These NTN-enabled applications exhibit fundamentally different characteristics from their terrestrial counterparts.
First, NTN comprises multi-layer orbits offering distinct coverage-latency tradeoffs~\cite{Azari2022Evolution,Jamshed2025NonTerrestrial}.
Second, NTN suffers severe propagation loss due to a transmission distance of thousands of kilometers, with atmospheric attenuation including rain fade at Ku/Ka bands and ionospheric scintillation at L/S bands.
Third, NTN experiences significant Doppler shifts as satellites move at 7\,km/s, making signal tracking highly challenging.
These new characteristics result in challenges across protocol layers:
\begin{itemize}
\item \textbf{Physical Layer}: Radio resource management faces severe interference among overlapping satellite beams. Accurate CSI estimation is challenging due to rapid satellite movement and long feedback delays, often resulting in outdated channel knowledge~\cite{2024Medina3GPP}.

\item \textbf{MAC Layer}: Random access protocols should handle massive connection requests from ground terminals with limited collision detection. Robot scheduling faces additional complexity due to time-varying intermittent connectivity~\cite{Azari2022Evolution}.

\item \textbf{Network Layer}: Mobility management requires frequent handovers with visibility windows lasting only minutes. Signal strengths fluctuate due to elevation angles and atmospheric losses~\cite{Ma2024Satellite}.

\item \textbf{Application Layer}: Embodied tasks involve non-uniform contributions across agents and regions. Existing approaches that target throughput or coverage maximization fail to capture end-to-end requirements of embodied tasks~\cite{Li2026MemoryCentric}.
\end{itemize}

\begin{table*}[!t]
\centering
\caption{Comparison of Terrestrial, NTN, and MemNTN Paradigms for EI}
\label{tab:paradigm_comparison}
\small
\scalebox{1.0}{
\begin{tabular}{|l|c|c|c|}
\hline
\textbf{Feature} & \textbf{Terrestrial-EI} & \textbf{NTN-EI} & \textbf{MemNTN-EI} \\
\hline
Network topology & Fixed (static stations) & Dynamic (moving satellites) & Dynamic + memory-enhanced \\
\hline
Link stability & \checkmark Stable & $\times$ Highly unstable & \checkmark Predictable via memory \\
\hline
Coverage range & Regional (km$^2$) & Global (planet-wide) & Global (planet-wide) \\
\hline
Resource constraints & + Moderate & +++ Severe & ++ Memory-optimized \\
\hline
Decision paradigm & Stateless/Cloud-assisted & Stateless & Memory-driven \\
\hline
Latency profile & 10--50 ms & 100--500 ms (variable) & 100--200 ms (prediction-aided) \\
\hline
Handover frequency & Low (minutes--hours) & Very high (seconds--minutes) & Optimized via trajectory memory \\
\hline
Adaptability & Reactive to local changes & Reactive to link changes & Proactive via experience \\
\hline
CSI acquisition & Direct measurement & Delayed/Outdated & Predicted from CSI memory \\
\hline
Computation distribution & Cloud/Edge only & Edge/Satellite & Distributed with memory hierarchy \\
\hline
Fault tolerance & High redundancy & Limited redundancy & Enhanced via memory redundancy \\
\hline
Scalability & Limited by BS density & Limited by satellite count & Enhanced via memory compression \\
\hline
\end{tabular}
}
\end{table*}

\subsection{Opportunity: From Stateless to Memory-Native NTN}

The MemNTN paradigm represents a fundamental shift by treating memory as a central element in network design.
As shown in Fig.~\ref{fig2}, memory in this framework follows a complete life cycle consisting of five stages as follows:
\begin{itemize}
    \item[(i)] \emph{Memory Acquisition}: Raw data is collected from sensors for physical memory or from network measurements for digital memory.
    \item[(ii)] \emph{Memory Compression}: Data is processed and encoded into appropriate representations (e.g., point clouds, feature vectors) and stored in memory banks located at appropriate network nodes (e.g., satellites, ground stations, edge robots).
    \item[(iii)] \emph{Memory Valuation}: Each memory item is assigned a score using the memory value model, reflecting the memory's usefulness for future tasks.
    \item[(iv)] \emph{Memory Update}: Guided by the updated memory values, new observations are fused with existing memories, while outdated memories are gradually forgotten.
    \item[(v)] \emph{Memory Utilization}: When making decisions such as routing, scheduling, or beamforming, relevant memories are retrieved based on their value scores and used to inform the decision-making process.
\end{itemize}
The above pipeline ensures that memory remains relevant and useful over time, adapting to changes in both physical and network conditions.

\subsection{Benefits Offered by Memory-Native NTN}

The proposed paradigm is particularly important to NTN, since \emph{the more uncertain the network, the more valuable memory becomes}. Specifically, the benefits extend across layers:
\begin{itemize}
\item \textbf{Channel Memory (Physical Layer)}: Historical CSI patterns capture spatiotemporal correlation of channel states, enabling prediction of future link qualities and compensating for outdated feedback caused by long propagation delays and rapid satellite movement.

\item \textbf{Access Memory (MAC Layer)}:
Collision histories from previous connection attempts can be leveraged to optimize random access protocols, reducing failures under massive connection requests with limited collision detection capabilities.

\item \textbf{Trajectory Memory (Network Layer)}:
Orbital and robotic trajectory information enables visibility window prediction, proactive routing, handover preparation, and network resource prefetch.

\item \textbf{Task Memory (Application Layer)}:
Task semantics and memories across agents and regions enable end-to-end task-aware memory compression, value prioritization, and resource allocation, instead of sub-module optimization.
\end{itemize}

Based on the above discussions, the MemNTN framework provides three key capabilities: (i) \emph{predictive} capabilities that enable anticipation of future link states and proactive resource allocation; (ii) \emph{experience reuse} that accelerates decision-making by leveraging learned patterns; and (iii) \emph{cross-layer intelligence} that provides a unified abstraction accessible across protocol layers.
These capabilities empower MemNTN to fully exploit long-horizon cross-layer memory rather than instantaneous task-layer semantic importance in existing semantic/goal-oriented communication \cite{Zheng2025Semantic}.
A comprehensive comparison of terrestrial EI, NTN-EI, and MemNTN-EI is provided in Table~\ref{tab:paradigm_comparison}.

\subsection{Distributed Memory Fusion}

In general, the construction and encoding of memories at individual agents might be missing, noisy, or erroneous due to sensor limitations, environmental complexity, or communication dropouts. Properly leveraging redundant observations from other agents can help mitigate such uncertainties and improve memory fidelity. To achieve a globally consistent memory across the network, distributed memory fusion is required, whereby multiple agents collectively contribute to and benefit from shared memory, as shown in Fig.~\ref{fig2}. Below we take physical memory as an example for illustrating the fusion process.

\subsubsection{Local Memory Construction}
Each NTN node constructs its local memory through autonomous operation:
\begin{itemize}
\item \textbf{Robots}: Execute simultaneous localization and mapping (SLAM) to build local poses and maps, extract visual features using vision-language models (VLMs), and compress data for transmission.
\item \textbf{Ground Stations}: Act as memory hubs with sufficient storage and computing resources for large language model (LLM) inference.
\item \textbf{Satellites}: Maintain orbital position awareness, collect wide-area observations, and build global context.
\end{itemize}

\subsubsection{Global Memory Fusion}
Robot memory chunks are first transmitted and fused at memory hubs. The fused local memory databases are then compressed and uploaded to satellites during the scheduled visibility windows, thereby constructing a consistent global memory lake across the NTN.
The global memory fusion consists of the following key functionalities.

\begin{itemize}
\item \textbf{Time Synchronization}:
It integrates observations with heterogeneous latencies and timestamps into a unified sequence aligned in the time domain.

\item \textbf{Spatial alignment}:
It transforms observations from different viewpoints to a common coordinate frame via extrinsic calibration (i.e., rotation and translation matrices).

\item \textbf{Redundancy management}: It addresses contradictory observations by data-level, feature-level, or output-level fusion techniques based on deep neural networks, while automatically identifying and removing redundant information arising from observation overlap.

\end{itemize}

\begin{figure*}[!t]
    \centering
    \includegraphics[width=0.98\textwidth]{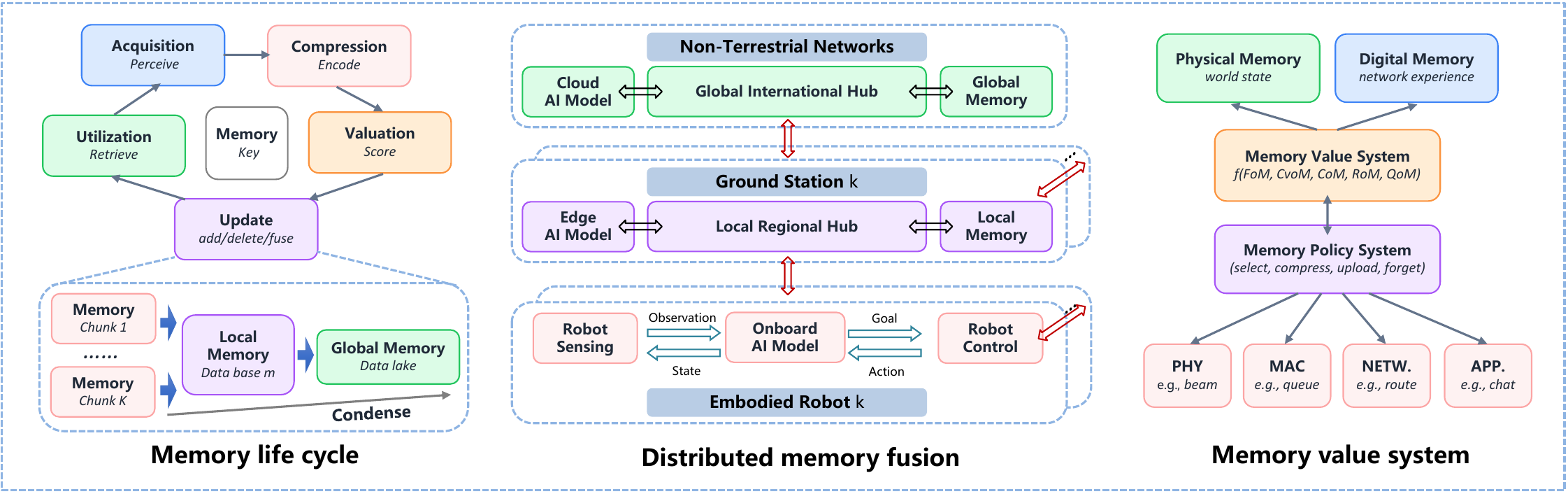}
    \caption{System framework of MemNTN consisting of memory life cycle, memory fusion, and memory valuation.}
    \label{fig2}
\end{figure*}

\section{Dichotomous Memory Representation}

\begin{figure}[!t]
    \centering
    \includegraphics[width=0.48\textwidth]{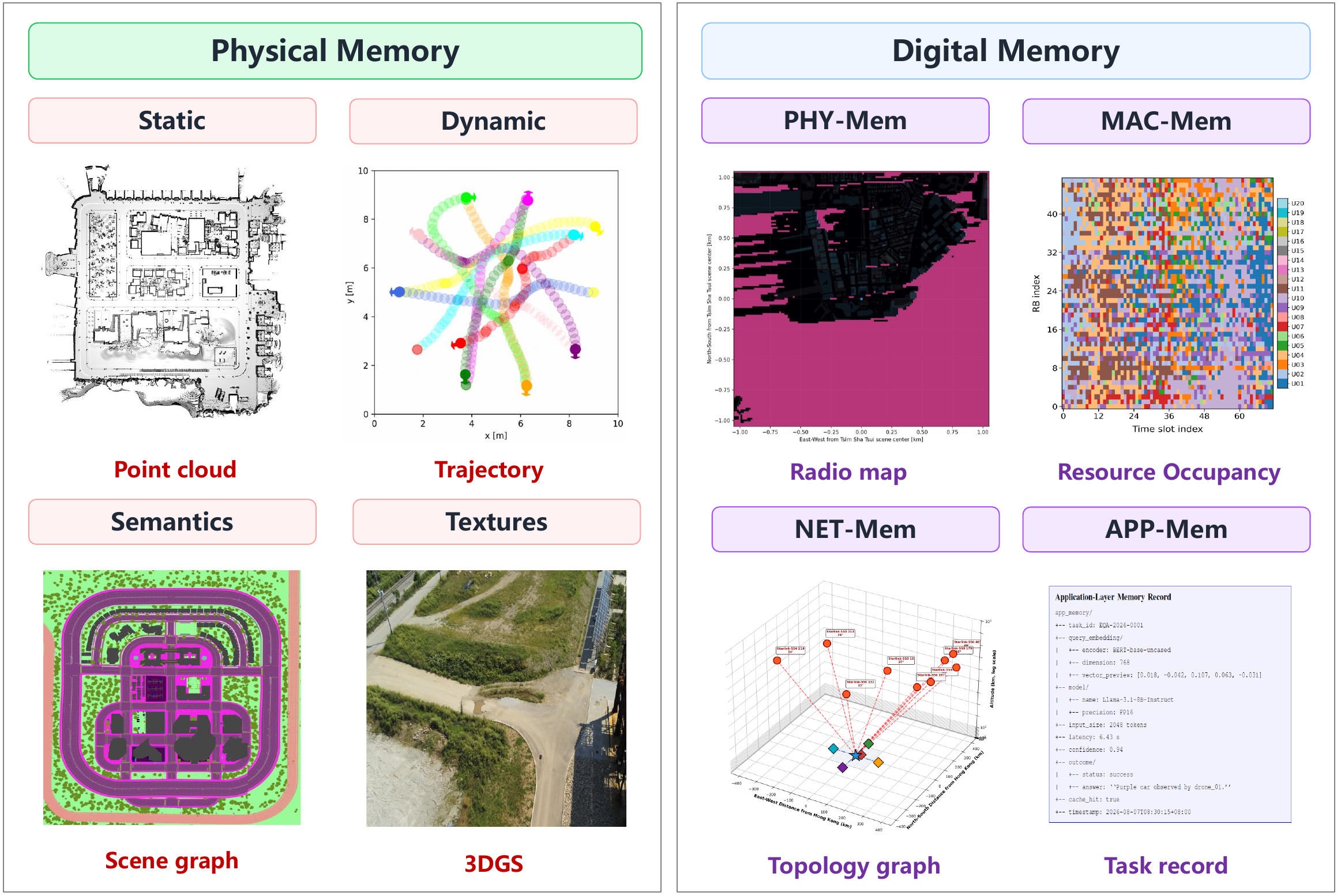}
    \caption{
    Dichotomous memory representation of MemNTN.}
    \label{fig3}
\end{figure}

As shown in Fig.~\ref{fig3}, the channel, access, trajectory, and task memories can be categorized into two fundamentally different types of memory.
In particular, physical memory represents the environment, including static geometry, dynamic poses and trajectories, semantic descriptions, object categories, and spatial relationships. It is acquired through multi-modal sensing and processed by object detection, semantic segmentation, scene understanding, and 3D reconstruction.
On the other hand, digital memory represents network operational experience, including radio maps, spectrum occupancy, scheduling records, link-quality trends, and handover outcomes. It is acquired from PHY/MAC/NET/APP measurements, control signaling, and radio access network interfaces.

\subsection{Physical Memory Representations}

Physical memory requires multiple representations to support different tasks.

\subsubsection{Statics} 3D point clouds provide high-precision geometry for localization and mapping but are costly to transmit over NTN links.
Compact representations such as grid and occupancy maps compress the data while preserving geometry granularity.

\subsubsection{Dynamics} Spatio-temporal memory captures agent positions, orientations, trajectories, velocities, and historical paths with timestamps. It supports motion prediction, collision avoidance, and multi-agent coordination, and can be compressed using curve fitting, probabilistic modeling, or keyframe sampling.

\subsubsection{Semantics}
Multi-modal data can be converted to scene descriptions using VLMs.
These semantics can then be leveraged by LLMs for information retrieval or chain-of-thought reasoning.
Semantics can also be parsed into spatial scene graphs for advanced embodied operations such as autonomous exploration and object search.

\subsubsection{Textures}
By representing scenes with a collection of learnable Gaussian primitives, 3D Gaussian Splatting (3DGS) achieves real-time rendering while preserving fine-grained appearance details.
It provides a compact representation of 3D environments, enabling novel view synthesis, landmark identification, and traversability estimation for embodied robots.

\subsection{Digital Memory Representations}

Digital memory requires multiple representations to support different network operations.

\subsubsection{Radio Tensors}
Radio tensors are multi-dimensional arrays representing channel states across time, frequency, and spatial dimensions.
They capture channel gains, Doppler shifts, phase
information, and interference patterns experienced by NTN
links.
These tensors can be acquired through pilot signal measurements, ray tracing, or machine learning methods, and
compressed by tensor decomposition.

\subsubsection{Resource Occupancy Grid Maps}
Resource occupancy maps encode spectrum patterns from MAC logs for predictive resource allocation and contention avoidance.
Matrix completion techniques can fill in unobserved map entries, enabling inference about spectrum availability during monitoring gaps. To facilitate sharing of resource maps, it is possible to compress dense occupancy maps into sparse counterparts.

\subsubsection{Network Topology Graphs} Network topology graphs represent satellites, ground stations, and robots as nodes, with radio, optical inter-satellite, and backhaul links as edges. Edge attributes include quality, capacity, latency, and resource availability. These graphs are built from routing tables and link-state advertisements, which support proactive handover under satellite motion.

\subsubsection{Task Records}
Application-layer task records store the identifier, query embedding, model, precision, input size, latency, confidence, outcome, cache-hit status, and timestamp. 
These records are automatically stored and retrieved by LLM agents, which support EI-native task orchestrations.

\section{Memory Value and Policy System}

\subsection{Multi-Dimensional Value Function}

Different memory items have varying relevance to current tasks, requiring the system to determine what to remember for decision-making.
Furthermore, the value of memory changes over time as conditions evolve, necessitating decisions about when to forget outdated information.
Consequently, the memory value model serves as a key component enabling effective memory usage in MemNTN.
By assessing the value of physical memory (e.g., importance of a geographic region) and digital memory (e.g., quality of a link), the system can make coherent decisions that respect both environmental and network constraints.

As shown in Fig.~\ref{fig2}, we first normalize the five dimensions to a common range and define the value of a memory item $\mathcal{M}$ as
\begin{align}
\mathrm{Value}(\mathcal{M}) &=
\mathbf{w}^{T}
\big[\mathrm{FoM}, \mathrm{CvoM}, -\mathrm{CoM}, \mathrm{RoM}, \mathrm{QoM}\big]^T,
\nonumber \\
\mathbf{w} &\in
  \left\{
  \mathbf{w}\in\mathbb{R}_{+}^{5}
  \,\middle|\,
  \mathbf{1}^{T}\mathbf{w}=1
  \right\}.
\nonumber
\end{align}
Here, $\mathbf{w}$ represents the relative importance assigned to the five normalized dimensions, and the negative sign makes CoM a cost term. The five dimensions are defined as follows.
\begin{itemize}
\item \textbf{Freshness of Memory (\text{FoM})}: How recently was the memory last verified or updated? Fresh memories are more likely to reflect current conditions.

\item \textbf{Coverage of Memory (\text{CvoM})}: What spatial, temporal, or device range does the memory apply to? Broadly applicable memories have higher value.

\item \textbf{Cost of Memory (\text{CoM})}: What are the storage, transmission, and retrieval costs? High-cost memories must provide commensurate benefits.

\item \textbf{Reusability of Memory (\text{RoM})}: Can the memory be applied across different scenarios, nodes, or tasks? Reusable memories have higher long-term value.

\item \textbf{Quality of Memory (\text{QoM})}: How much does a memory reduce uncertainty about current or future states? Memories with high information gain or error reduction have higher value.
\end{itemize}

Note that $\mathbf{w}$ should be configured according to the target task. For example, a delay-sensitive mission may assign a larger weight to FoM, a mapping task may emphasize CvoM, a resource-constrained deployment may place a larger weight on the negative CoM term, and a knowledge-reuse application may emphasize RoM. The weights can be selected using cross validation over task datasets.

It can be seen that memory value accounts not only for semantic relevance, but also for temporal validity, novelty, representativeness, storage/transmission cost, and future reusability. Consequently, observations with similar semantic embeddings can have substantially different memory values.

\subsection{Computational Valuation Methods}

Each value dimension can be instantiated using a metric-specific computational model.
\begin{itemize}
    \item First, the FoM can be quantified by age of information (AoI), which captures both memory generation time and delivery delay.
    \item  Second, the CvoM can be computed from geometric visibility and communication reachability by combining sensing geometry (e.g., field-of-view overlap and spatial footprint) with propagation-aware connectivity (e.g., path loss and blockage probability).
    \item Third, the CoM can be estimated from storage, transmission, and retrieval overhead using analytical models (e.g., memory size, compression ratio) or empirical regression models fitted from logs (e.g., latency, energy).
    \item Fourth, the RoM can be measured by the mutual information between different items, where higher mutual information indicates stronger cross-scenario, cross-node, and cross-task reuse value.
    \item Lastly, direct QoM computation is difficult, as it depends on task semantics, context, and evolving uncertainty. To address this, we adopt an agentic QoM model in which an agent dynamically selects and composes valuation tools, as detailed in the next subsection.
\end{itemize}

\subsection{Agentic Quality of Memory Model}

We adopt an agentic QoM model based on a generative adversarial exam (GAE) workflow~\cite{Li2026MemoryCentric}.
The approach transforms the computation of QoM into a memory-retrieval question answering process. In our setting, this memory can be either \emph{physical memory} (e.g., scene layout, objects, trajectories, semantic descriptions) or \emph{digital memory} (e.g., channel condition history, access outcomes, handover records, resource occupancy). Representative question-answer pairs are as follows.
\begin{itemize}
\item Question on physical memory: Which side of the collapsed building is passable for a ground robot? Answer: The east-side remains open, while the west side is blocked.
\item Question on digital memory: Which satellite provides the most stable uplink for this area? Answer: Satellite S238 provided the highest stability during the last visibility window.
\end{itemize}
Then we can estimate the incremental value of each node's memory by testing whether the current global memory can answer questions generated from that node's own observations.
If the current global memory fails these questions, the node is likely to contribute novel and valuable memory.
However, if the global memory already answers them well, the added value is limited.
Consequently, the GAE pipeline operates according to the following three stages.
\begin{itemize}
\item \textbf{Pilot Upload}: Each node uploads a small, randomly sampled subset of local data. This keeps overhead low while preserving enough semantic diversity for valuation.
\item \textbf{Exam Generation}: The server converts pilot data into pilot memory and generates question-answer pairs from that memory using LLMs. The generated question–answer pairs are grounded in the pilot observations and are subsequently validated against the source data.
\item \textbf{Practice Test}: The server tests existing global memory against these generated questions. The resulting score reflects how much of the node's information is already covered by current memory.
\end{itemize}

This exam-style valuation captures memory novelty. A lower test score indicates stronger novelty and higher potential contribution to global memory quality, while a higher score indicates redundancy.
Notably, the valuation is performed using pilot data, thereby keeping the communication overhead minimal.
To control computation overhead, we use iterative memory retrieval rather than feeding the entire global memory context into the LLM.
For each question, the LLM agent first selects a text-, position-, or time-based retrieval tool according to the question and the currently accumulated context. The database then returns the nearest memory chunks, i.e., Top-$5$ for text retrieval, Top-$4$ for position retrieval, and Top-$4$ for time retrieval.
The LLM then appends their captions, timestamps, and poses to the working context. The stopping criterion is operationally defined as either (i) the agent produces an answer without retrieval request or (ii) a three-call limit is reached.
The adopted limits bound each question to at most three database queries and 15 retrieved chunks.
This reduces context length, improves scalability, and keeps valuation latency practical for NTN settings with distributed agents.
Overall, the agentic QoM model provides a unified framework for computing the semantic usefulness for EI tasks, enabling the network to make memory-native decisions.

\subsection{Value-Driven Policy}

After memory quality is estimated, the network applies a value-driven policy to determine upload priority, link scheduling, and resource allocation.
In each scheduling round, the policy selects actions that maximize expected task gain under NTN infrastructure constraints.
This implies that the policy should jointly consider physical and digital memories.
Specifically, physical memory provides task utility signals such as scene criticality, target relevance, and geographic coverage gain, while digital memory provides system feasibility signals such as channel stability, interference, queue state, handover risk, and delivery success.

To achieve this objective, memory value is propagated across all NTN protocol layers, governing memory selection at the application layer, adjusting handover preferences at the network layer, configuring queue priorities at the access layer, and controlling beamforming designs at the physical layer.
High value memories are preferentially routed through stable paths and reliable links whenever feasible, whereas low value uploads are delayed, compressed, or accommodated using residual capacity.
Furthermore, a fairness aware coordination mechanism reserves periodic transmission opportunities for nodes with weak channel conditions, thereby preventing long term starvation.
Consequently, NTN operation is transformed from stateless to memory native optimization, aligning network conditions with physical task performance.

\section{Case Study}

\begin{figure*}[!t]
    \centering
    \includegraphics[width=0.98\textwidth]{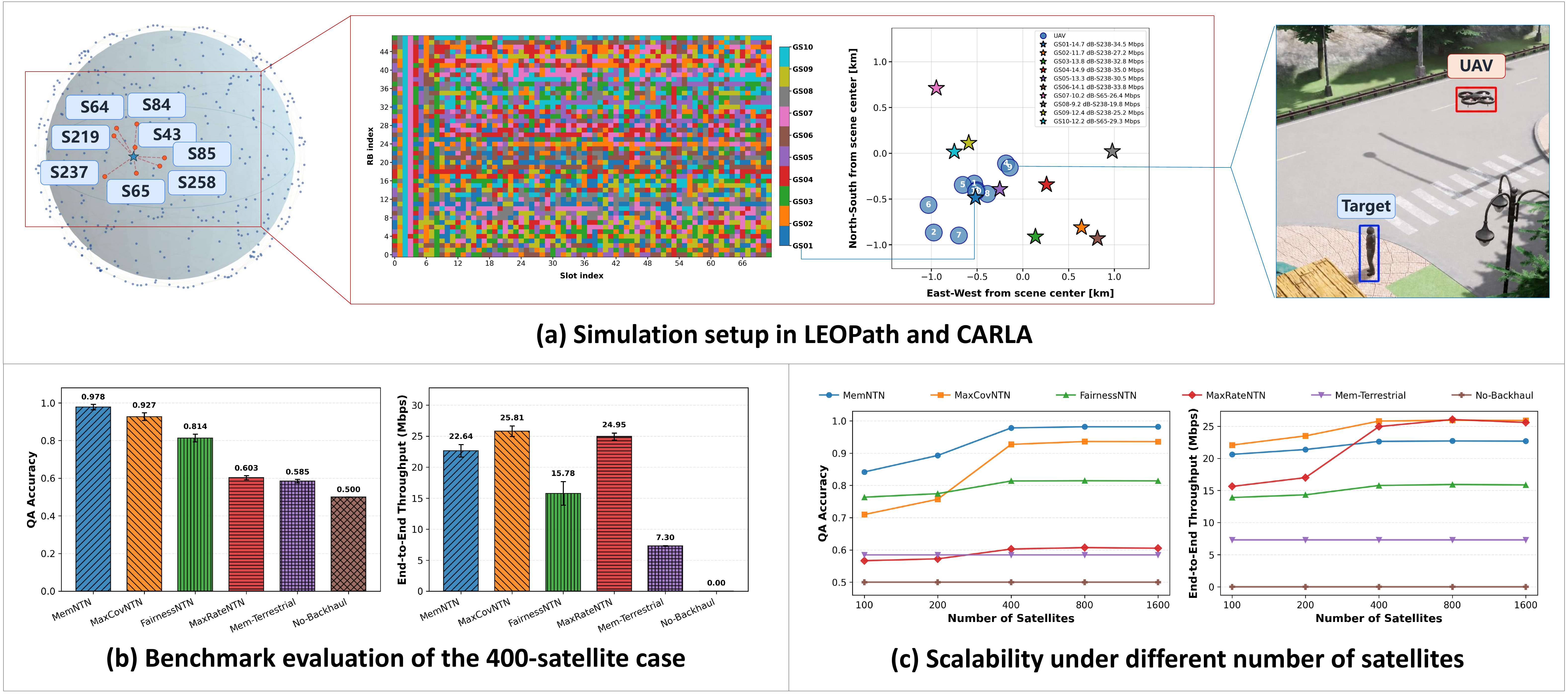}
    \caption{Verification of MemNTN in SEQA tasks. (a) Simulation settings in LEOPath and CARLA platforms; (b) Performance evaluation of the 400-satellite case; (c) Task accuracy and end-to-end throughput versus the number of satellites.
    }
    \label{fig4}
\end{figure*}

We evaluate the effectiveness of MemNTN in satellite embodied question answering (SEQA) tasks on a Linux workstation with an NVIDIA RTX 3090 Ti.
To emulate the NTN communication environment, we integrate the CARLA platform\footnote{\url{https://github.com/carla-simulator/carla}} with AirSim\footnote{\url{https://github.com/louiszengCN/CarlaAir}}, LEOPath\footnote{\url{https://github.com/Fundacio-i2CAT/LEOPath/}}, and OpenNTN\footnote{\url{https://github.com/ant-uni-bremen/OpenNTN}}.
AirSim provides the unmanned aerial vehicle (UAV) motion dynamics.
LEOPath provides low earth orbit (LEO) satellite trajectories, visibility windows, elevation angles, propagation delays, and handover timing between the ground station and satellites.
OpenNTN provides the NTN channel abstraction, including path loss, Doppler dynamics, atmospheric attenuation, and time-varying link quality.
During each simulation run, CARLA produces physical memory from UAV sensing, LEOPath determines whether a satellite is visible to the ground station, and OpenNTN maps the satellite geometry into the available backhaul throughput. The ground station acts as the memory hub, which receives UAV memories through the local access links, selects memories using the value model, and forwards selected memory updates through the NTN backhaul to the remote command center.
In the controlled optimization, FoM, CvoM, CoM, and RoM are held constant across UAVs and the memory-value ordering depends only on QoM.

\subsection{Simulation Setup}

\begin{table*}[t]
\centering
\caption{Simulation Parameters for the Satellite Embodied Question Answering Case Study}
\label{tab:sim_params}
\resizebox{\textwidth}{!}{
\renewcommand{\arraystretch}{1.2}
\small
\begin{tabular}{@{}|c|c|c|c|@{}}
\hline
\rowcolor{gray!25}
\textbf{LEO Constellation} & \textbf{Satellite Backhaul} & \textbf{UAV Access Link} & \textbf{CARLA Setup} \\
\hline
\begin{tabular}{@{}l@{\quad}l@{}}
Satellite counts: & $400$ ($20\times20$) \\
Altitude: & $\approx509$\,km \\
Inclination: & $53^{\circ}$ \\
Orbital period: & $94.8$\,min \\
Visible satellites: & $8$ \\
Min elevation: & $10^{\circ}$ \\
Ground station: & Hong Kong \\
Mean motion: & $15.19$/day
\end{tabular}
&
\begin{tabular}{@{}l@{\quad}l@{}}
Uplink: & HK GS01 $\!\rightarrow\!$ S238 \\
Downlink: & Istanbul S119 $\!\rightarrow\!$ GW \\
Uplink frequency: & $30$ GHz \\
Multiplex: & OFDMA \\
Scheduler: & PF \\
Resource grid: & $48$\,RB $\times 72$\,slots \\
GS$01$ EIRP: & $58$\,dBm \\
Satellite G/T: & $13$\,dB/K
\end{tabular}
&
\begin{tabular}{@{}l@{\quad}l@{}}
UAV counts: & $10$ \\
Serving memory hub: & GS$01$ \\
Antennas: & $256$ \\
Multiplex: & MIMO \\
Sum power: & $300$\,mW \\
Noise: & $-70$\,dBm \\
Bandwidth: & $10$\,MHz \\
Access duration: & $600$\,s
\end{tabular}
&
\begin{tabular}{@{}l@{\quad}l@{}}
UAV counts: & $10$ \\
Target objects: & $10$ \\
Map: & Town04 \\
Num. of QAs: & $30$/run\\
Perception time: & $30$\,s \\
Frame rate: & $35$\,FPS \\
Frames per UAV: & $1050$ \\
Total frames/run: & $10500$
\end{tabular}
\\
\hline
\end{tabular}
}
\end{table*}

Table~\ref{tab:sim_params} and Fig.~\ref{fig4}a summarize the simulation parameters and illustrate the system configuration, respectively. The LEO constellation follows a Walker star configuration with 400 satellites at an altitude of approximately 509\,km and an inclination of 53$^{\circ}$, resulting in an orbital period of 94.8\,minutes and a mean motion of 15.19\,revolutions per day, consistent with common LEO-NTN modeling assumptions~\cite{Azari2022Evolution,Ma2024Satellite}. We simulate ten backhaul nodes GS01--GS10 in Hong Kong. The GS01 acts as the active memory hub serving ten UAVs, while the remaining nodes model competing terminals.\footnote{For the multi-hub scenario, under sufficiently effective interference suppression and coordinated beamforming, the problem can be decomposed into parallel single-hub subproblems.}

End-to-end memory delivery follows UAV $\rightarrow$ GS01 $\rightarrow$ S238 $
\xrightarrow{\text{ISL route}}$ S119 $\rightarrow$ Istanbul GW. The ISL route is assumed to have sufficient capacity and
therefore does not constitute a bottleneck. The effective satellite-backhaul rate is the minimum of the GS01--S238 uplink rate
and the S119--Istanbul GW downlink rate. Both links use OFDMA and a proportional-fair scheduler over a grid
of 48 resource blocks and 72 time slots~\cite{2024Medina3GPP,Liu2024TransmitPower}. Each resource block contains 12
subcarriers with a subcarrier spacing of 120\,kHz. 
The downlink rate consistently exceeds the uplink rate, and the GS01--S238 uplink determines the reported backhaul capacity.

The CARLA simulation runs on the Town04 map. Each UAV is equipped with a camera operating at 35~FPS, with each image compressed to 200\,KB for transmission. The CARLA perception phase lasts 30\,s, yielding 1050 frames per UAV. 
Each run generates 10 target objects. For each target, the questions involve $3$ types:
(i) whether a target object exists; (ii) where the target object is located; and (iii) which UAV observes the target.
We use \textit{Qwen3-VL-8B} for image captioning, \textit{mxbai-embed-large-v1} for text embedding, \textit{Milvus} for memory database construction, and \textit{Qwen3-8B} for question answering~\cite{Li2026MemoryCentric,Zheng2025Semantic}.\footnote{
The captioning latency of \textit{Qwen3-VL-8B} is $4.2$\,s per 5 images, and the inference latency of \textit{Qwen3-8B} is $6.43$\,s per QA.
}

\subsection{Benchmark Results}

We evaluate the system using SEQA accuracy and end-to-end throughput at the remote command center.
We compare the proposed MemNTN scheme with representative baselines: (1) \textbf{MaxRate}~\cite{Kao2025Analytical}, which prioritizes high-throughput links; (2) \textbf{MaxCov}~\cite{Li2026MemoryCentric}, which prioritizes sensing coverage; (3) \textbf{Fairness}~\cite{Liu2024TransmitPower},which allocates resources among UAVs according to the max-min fairness criterion; (4) \textbf{Mem-Terrestrial} \cite{Li2026MemoryCentric}, which uses memory-based scheduling with terrestrial backhaul (i.e., fiber-connected ground station with 1--20 Mbps capacity); and (5) \textbf{Mem-No-Backhaul}~\cite{Li2025Embodied}, which represents a completely disconnected scenario.

Fig.~\ref{fig4}b summarizes the numerical results across 100 simulation runs.
MemNTN achieves the highest remote SEQA accuracy among all compared schemes, achieving relative accuracy improvements of 62.3\%, 5.5\%, 20.2\%, and 67.2\% over MaxRate, MaxCov, Fairness, and Mem-Terrestrial, respectively.
This confirms that the final task performance is not determined solely by the amount of transmitted data, but by whether the delivered memories contain novel and task-relevant information. MaxRate obtains high communication throughput but forwards visually redundant observations, while Mem-Terrestrial suffers from congested terrestrial access that limits effective memory delivery.

The results demonstrate the benefit of jointly considering memory relevance and network conditions in the considered SEQA setting.
Simply maximizing the local wireless rate is insufficient, because the memory may be of low value. Conversely, simply selecting the most semantically valuable UAV memory may also fail if that UAV has a weak access link or if the satellite forwarding window is too short.
MemNTN addresses this problem by jointly considering physical memory value and digital memory state. Physical memory indicates whether a UAV observes novel objects or important regions, while digital memory records access-link quality, satellite visibility, backhaul congestion, and recent delivery success.

\subsection{Scalability Analysis}

Fig.~\ref{fig4}c shows how SEQA accuracy and end-to-end throughput scale with the number of satellites. We evaluate configurations with 100, 200, 400, 800, and 1600 satellites, keeping all other parameters fixed. The QA accuracy of MemNTN improves from 84.2\% with 100 satellites to 98.2\% with 1600 satellites. MaxCov approaches MemNTN performance at larger constellation sizes because abundant backhaul reduces the need for selective transmission.

For MemNTN, the QA accuracies are $98.17\%$ and $98.16\%$ in the $800$- and $1600$-satellite cases, respectively, demonstrating that performance saturates beyond 800 satellites. The backhaul capacity increases from $22.51$\,Mbps with 100 satellites to $36.70$\,Mbps with 800 satellites, whereas end-to-end throughput increases only from $20.62$ to $22.71$\,Mbps. 
This shows that the satellite backhaul is no longer the bottleneck and further constellation densification yields diminishing gains.

The 1600-satellite case is marginally lower than the 800-satellite case. 
This small difference suggests that the effective satellite-backhaul rate also depends on satellite geometry rather than solely on constellation size.

\section{Conclusion and Open Directions}

This paper proposes MemNTN, a memory-native architecture for supporting embodied intelligence over non-terrestrial networks. Instead of treating each transmission as an isolated event, MemNTN organizes physical memory of the environment and digital memory of network experience into a shared life cycle of acquisition, compression, valuation, update, and utilization. Through this dichotomous memory representation, NTN nodes can reuse past observations, predict future link and task conditions, and coordinate decisions across the physical, access, network, and application layers. The SEQA case study shows that task performance depends not only on delivered data volume, but also on whether the delivered memory is novel, reliable, and relevant to downstream reasoning.

Two important future directions are listed as follows. First, distributed memory consistency and standardization should be jointly addressed through lightweight synchronization, conflict-resolution mechanisms, and interoperable interfaces that allow memory value to be exchanged across satellites, ground stations, robots, and emerging 6G control planes. Second, memory security and privacy deserve particular attention because MemNTN stores both sensitive environmental observations and operational network traces. Future systems should defend against memory poisoning, unauthorized retrieval, and inference attacks that expose mission locations or user intents.

\bibliographystyle{IEEEtran}
\bibliography{ref}

\end{document}